\documentclass{article}

\usepackage{microtype}
\usepackage{graphicx}
\usepackage{subfigure}
\usepackage{float}
\usepackage{booktabs} 
\usepackage{hyperref}
\usepackage[normalem]{ulem}
\usepackage{comment}

\usepackage[accepted]{pre-training}

\icmltitlerunning{Knowledge Distillation for Efficient Sequences of Training Runs}

\begin{document}

\twocolumn[
\icmltitle{Knowledge Distillation for Efficient Sequences of Training Runs}




\begin{icmlauthorlist}
\icmlauthor{Xingyu Liu}{harvard}
\icmlauthor{Alex Leonardi}{harvard}
\icmlauthor{Lu Yu}{harvard}
\icmlauthor{Chris Gilmer-Hill}{harvard}
\icmlauthor{Matthew Leavitt}{MosaicML}
\icmlauthor{Jonathan Frankle}{MosaicML}
\end{icmlauthorlist}

\icmlaffiliation{harvard}{Harvard University, Cambridge, USA.}
\icmlaffiliation{MosaicML}{MosaicML, San Francisco, USA.}

\icmlcorrespondingauthor{Jonathan Frankle}{jonathan@mosaicml.com}
\icmlcorrespondingauthor{Matthew Leavitt}{matthew@mosaicml.com}

\icmlkeywords{Machine Learning, ICML}

\vskip 0.3in
]



\printAffiliationsAndNotice{}  

\newif\ifshow
\showtrue

\ifshow
\newcommand{\jonathan}[1]{\textcolor{blue}{[\textbf{Jonathan:} #1]}}
\newcommand{\matthew}[1]{\textcolor{green}{[\textbf{Matthew:} #1]}}
\newcommand{\todo}[1]{\textcolor{red}{[TO DO: #1]}}
\else
\newcommand{\ari}[1]{}
\newcommand{\mat}[1]{}
\newcommand{\todo}[1]{}
\fi

\begin{abstract}

In many practical scenarios---like hyperparameter search or continual retraining with new data---related training runs are performed many times in sequence.
Current practice is to train each of these models independently from scratch.
We study the problem of exploiting the computation invested in previous runs to reduce the cost of future runs using knowledge distillation (KD).
We find that augmenting future runs with KD from previous runs dramatically reduces the time necessary to train these models, even taking into account the overhead of KD.
We improve on these results with two strategies that reduce the overhead of KD by 80-90\% with minimal effect on accuracy and vast pareto-improvements in overall cost.
We conclude that KD is a promising avenue for reducing the cost of the expensive preparatory work that precedes training final models in practice.

\end{abstract}

\section{Introduction}
\label{intro}


In the real world, there are many scenarios in which model training must be performed repeatedly (e.g. manual model tweaking, explicit hyperparameter search, retraining on new data, etc.).
Doing so is expensive, especially since contemporary practice is to begin each new run from the same starting point, throwing away the work that was put into previous runs.
There is justification for this practice: starting from an already-trained model can make it difficult for the model's behavior to change in light of new conditions or data \citep{achille2018critical, ash}, undermining the benefits of beginning with learned features.
However, as training costs skyrocket into the millions of dollars, spending many times that amount developing the appropriate training procedure looks increasingly untenable.

In such scenarios, it would be useful to leverage information from prior training runs to train future models faster without explicitly reusing the models themselves.
In this paper, we develop and demonstrate the efficacy of one such approach: distillation from previous training runs.
\emph{Knowledge distillation} (KD)  is the process of training a model (known as the \emph{student}) to mimic the outputs of one or  more existing models (known as \emph{teachers}) rather than training on the labels associated with the data \citep{hinton2015distilling}.
KD has been used to condense ensembles of trained models into a single model \citep{you2017learning, chen2019two, park2020feature, asif2019ensemble} and to compress large trained models into smaller forms that are more amenable for inference.

In this paper, we evaluate KD as a way to convey information from prior training runs to future runs in a way that speeds up learning and reduces cost.
\textbf{The distinctive aspects of our problem setting are that (1) prior training runs are necessarily suboptimal} (otherwise, the current training run wouldn't be necessary) and \textbf{(2) minimizing the cost of KD is paramount} (otherwise, it wouldn't actually reduce the cost of future runs).

KD has several characteristics that make it a promising avenue for reducing the cost of future training runs.
It is softer than using the trained weights from a previous run.
It makes it possible to take advantage of multiple previous runs.
It is known to lead to effective learning; students can even reach a higher accuracy than their teachers \citep{allen2020towards}.
Finally, it can be regimented: not all teachers must necessarily be used on all steps, and some steps need not include any distillation.

In this paper, we study KD as a means to reduce the cost of future training runs in workflows of related runs like hyperparameter search.
We explore the design space for strategies that are both (1) effective even in the context of suboptimal teachers and (2) efficient enough to provide real reductions in training cost.
We focus on ResNet-56 on CIFAR-100, and we find that:



\textbf{Distillation makes future training runs cheaper.} Distilling from five identically trained teachers allows the student to reach the accuracy of any individual teacher in 70\% fewer steps or to reach accuracy 3.9 percentage points higher.
When the teachers result from hyperparameter search over learning rate (which leads to a diverse set of teachers, many of which are suboptimal), the student can still match the best teacher's accuracy in 66\% fewer steps and reach accuracy 3 percentage points higher when trained to completion.
Even accounting for the overhead of distillation, this is a pareto improvement over training the student from scratch.

\textbf{Distillation need not be applied on every step.}
Applying distillation on only 10\% of training steps still allows the student network to recover much of the aforementioned performance, mitigating the otherwise significant cost of performing distillation.
Strategically distributing those steps throughout training further improves this result.

\textbf{Not all teachers are necessary on all steps.}
These results also hold when we sample a small number of teachers---or even just one teacher---every time distillation is performed rather than using all teachers.
The model requires more steps and reaches slightly lower accuracy, but the reduction in distillation overhead is so dramatic that it is still a pareto improvement over distilling from all teachers or training normally.

We conclude that KD is a promising avenue for reducing the cost of a common and cripplingly expensive reality: developing and maintaining real-world models requires many related training runs, not just one.

\begin{table*}
 \begin{minipage}{0.35\textwidth}
 \makeatletter\def\@captype{table}\makeatother\caption{Accuracy at epoch 200 for teachers trained with the same hyperparameters and students distilled from 1-5 of those teachers.\vspace{1mm}}
 \centering
\begin{tabular}{lc} 
\toprule Model & Test Accuracy  \\
\midrule Teacher A0 & $65.6\%$ \\
     Teacher A1 & $67.4\%$ \\
     Teacher A2 & $68.3\%$ \\
     Teacher A3 & $66.0\%$ \\
     Teacher A4 & $65.8\%$ \\
KD (1 Teacher) & $69.2\%$ \\
KD (2 Teachers) & $71.2\%$ \\
KD (3 Teachers) & $72.0\%$ \\
KD (4 Teachers) & $72.2\%$ \\
KD (5 Teachers) & $72.2\%$ \\
\bottomrule
\end{tabular}
\label{tb:fix_t-acc}
 \end{minipage}
 \quad
  \begin{minipage}{0.63\textwidth}
\makeatletter\def\@captype{table}\makeatother\caption{Epochs necessary to reach the specified test accuracy for teachers trained with the same hyperparameters and students distilled from 1-5 of those teachers. \phantom{blah blah blah}\vspace{1mm}}
\begin{tabular}{lcccccc}
\toprule
Model & $65.0\%$ & $66.0\%$ & $67.0\%$ & $68.0\%$ & $69.0\%$ & $70.0\%$ \\
\midrule Teacher A0 & 96 & --- & --- & --- & --- & --- \\
Teacher A1 & 71 & 82 & 112 & --- & --- & --- \\
Teacher A2 & 57 & 67 & 92 & 124 & --- & --- \\
 Teacher A3 & 92 & 119 & --- & --- & --- & --- \\
 Teacher A4 & 89 & 130 & --- & --- & --- & --- \\
KD (1 Teacher) & 25 & 33	& 36 & 66 & 104 &--- \\
KD (2 Teachers) & 22 & 25 & 33 & 43 & 54 & 81 \\
KD (3 Teachers) & 19 & 21 & 27 & 35 & 48 & 64 \\
KD (4 Teachers) & 17 & 21 & 30 & 33 & 43 & 58 \\
KD (5 Teachers) & 17 & 22 & 27 & 33 & 42 & 59 \\
\bottomrule
\end{tabular}
\label{tb:fix_t-epoch}
\end{minipage}
\vspace{-5mm}
\end{table*}

\section{Experimental Framework}
\label{sec:exp_frame}

We focus on CIFAR-100 and ResNet-56. This benchmark is challenging enough to capture the properties of larger-scale settings while still permitting us to explore the design space of a paradigm in which each experiment requires multiple training runs using our limited computational resources.



In many real-world training workflows, model training must be performed repeatedly, resulting in multiple trained models of the same architecture. To investigate distillation in this context, for each experiment, we will train five ResNet-56 models on CIFAR-100 (teachers) that differ only in their random seed and hyperparameters.
We will then further train another ResNet-56 (student) model with knowledge distillation (the specifics of the distillation vary depending on the experiment). All teacher and student models are trained for 200 epochs, with a batch size of 128 and exponential learning-rate decay scheduler (decay factor $\gamma$ = 0.975). All student models are trained with same initial learning rate of 0.1.
When we train students, we perform three replicates of each training runs with different random initializations and present the average in our results.

We trained on AWS using P3.2xlarge GPU instances (NVIDIA V100s) with a pre-existing Deep Learning AMI. 



\begin{figure}
\centering
\includegraphics[width=\linewidth]{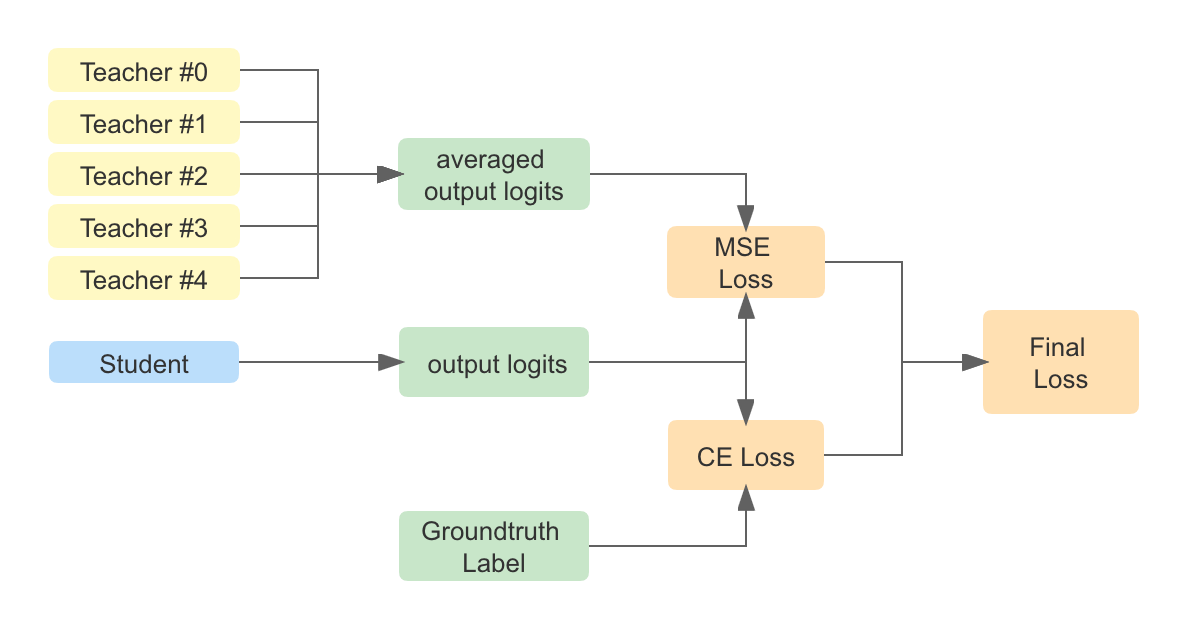}
\caption{KD Pipeline}
\label{fig: diagram}
\end{figure}

\section{The Efficacy of Knowledge Distillation}
\label{sec:kd_same}


In this section, we evaluate the efficacy of knowledge distillation as a tool for accelerating the training of student networks.
We find that it is highly effective at doing so, even when the teachers have different (and often suboptimal) accuracy as a result of hyperparameter search.

\textbf{Methodology.}
We train five different teachers using the setup from Section \ref{sec:exp_frame}.
We then train a student using one or more of these teachers via KD.
Concretely, to transfer knowledge from those teachers to the student, we use the averaged response from a set of $n$ teachers ($T$) as the supervision signal. Specifically, we adjust the loss for our student model to combine the cross-entropy ($CE$) between the student output logits ($\hat{y}_s$) and the ground truth ($y$) with the mean squared error ($MSE$) between the student output logits and an average teacher output logits ($\hat{y}_t$). Concretely, this modified loss is: $\mathcal{L}_{KD} = CE(\hat{y}_s, y) + MSE\left(\hat{y}_s, \frac{1}{n}\sum_{t\in T}\hat{y}_{t}\right)$

\textbf{KD with identical teachers.}
We begin by studying KD in settings where all teachers are trained with the same hyperparameters and differ only in random seed.
Although this setting is a simplification of our motivating real world scenarios, it is a proving ground for ideas and a chance to identify the best possible outcomes from using KD to accelerate future training runs.
If a strategy does not work in this simpler setting, we should not expect it to work when teachers are suboptimal.

We first train five teachers for this experiment using learning rate 0.1 and different random seeds.
We denote the teachers A0 through A4.
Tables \ref{tb:fix_t-acc} and \ref{tb:fix_t-epoch} show the final accuracy of each teacher and the speed at which the teachers learn; since they are trained with different random seeds, their performance varies slightly.




In Tables \ref{tb:fix_t-acc} and \ref{tb:fix_t-epoch}, we show the effect of distilling with the aforementioned loss when varying the number of teachers $T$ from 1 to 5. 
Increasing the number of teachers increases the final accuracy of the student and the time necessary  for it to reach any given level of accuracy, with diminishing returns as the number of teachers increases. With $\geq 3$ teachers, it can reach the same level of accuracy as the best individual teacher in less than 30\% of the epochs. When trained to completion, the student reaches an accuracy that is 3.9 percentage points higher than the best teacher.




\textbf{KD from diverse and suboptimal teachers.}
We next study the practical scenario in which the teachers differ.
We do so by simulating hyperparameter search over learning rate, varying the initial learning rate among the values 0.5, 0.2, 0.1, 0.05, and 0.01.
We denote these teachers as B0-B4, respectively.

As Table \ref{tb:diff_lr-acc} shows, these differences in learning rate affect the performance of the teachers appreciably.
Teacher B1 with learning rate 0.2 reaches the highest test accuracy (69.5\%), while Teacher B4 with learning rate 0.01 only reaches 62.1\%.
Tables \ref{tb:diff_lr-acc} and Table \ref{tb:diff_lr-epoch} show the result of performing knowledge distillation with these teachers.
Despite the diversity of the teachers, a student trained with all five teachers performs as well or better than using identically trained teachers and reaches 69\% accuracy in 33\% of the training time required by the best performing teacher.
 
 

To contextualize these results, we also study two baselines.
First, we consider a scenario where we use the weights of the best teacher (B1 with learning rate 0.2) as a starting point for training the new model (a baseline we refer to as \emph{fine-tuning}). With the initial 200 epochs, B1 has already saturated. When fintuning,  we observe that finetuning with lr=0.1 is better than small learning rate like 1e-4, this implies that the original teacher model may converge in a local minima and a large learning rate can help it jump out of the local minima.
Despite the fact that the new model is building on trained weights from the highest-performing teacher, it still reaches lower overall accuracy than the distilled model by 2 percentage points and learns more slowly once it begins modifying its initial weights.
This supports the findings in prior work that training may constrain the network's ability to learn in new ways \citep{achille2018critical, ash} and our contention that distillation offers a middle ground in which the new model can benefit from prior computation without being overly constrained by it.

Second, we consider a baseline where we allocate the additional computation needed by distillation in a different way.
The forward pass of training is approximately one third of the cost of training, meaning that distilling from five teachers increases the cost of training by to approximately $8 \over 3$ of its original value.
Instead of spending those additional resources on distillation, we also consider simply training a larger model that costs approximately twice as much to train: ResNet-110. We find that using the settings in the original paper \cite{he2016deep} can yield highest test accuracy: use learning rate of 0.01 to warm up in the first 400 iterations then switch back to 0.1, and divide the learning rate by 10 at 32k and 48k iterations. Nevertheless, this model still reach lower accuracy than the distilled model, and it trains more slowly.

\textbf{Summary.}
We conclude that, not only does KD allow the student model to train faster even in the real-world context of hyperparameter search with suboptimal teachers, but it outperforms two compelling alternatives.

 
 \begin{table*}
 \begin{minipage}{0.32\textwidth}
  \centering
     \makeatletter\def\@captype{table}\makeatother\caption{Accuracy at epoch 200 for teachers trained with different hyperparameters (learning rates), a student, and baselines.\vspace{1mm}}
\begin{tabular}{lc} 
\toprule Model & Test Accuracy  \\
\midrule Teacher B0 (0.5) & $69.3\%$ \\
Teacher B1 (0.2) & $69.5\%$ \\
Teacher B2 (0.1) & $66.1\%$ \\
Teacher B3 (0.05) & $65.1\%$ \\
Teacher B4 (0.01) & $62.1\%$ \\
KD (5 Teachers) & $72.5\%$ \\
Finetune Teacher B1  & $70.5\%$ \\
ResNet-110 & $71.4\%$ \\
\bottomrule
\end{tabular}
\label{tb:diff_lr-acc}
 \end{minipage}
 \quad
  \begin{minipage}{0.66\textwidth}
\makeatletter\def\@captype{table}\makeatother\caption{Epochs necessary to reach the specified test accuracy for teachers trained with different hyperparameters (learning rates), a student, and baselines.\\ \phantom{(blah blah blah blah blah blah blah blah blah} \vspace{1mm}}
\begin{tabular}{lcccccc}
\toprule
Model & $65.0\%$ & $66.0\%$ & $67.0\%$ & $68.0\%$ & $69.0\%$ & $70.0\%$ \\
\midrule 
Teacher B0 (0.5)  & 96& 102& 110 & 118  & 136  & ---   \\
Teacher B1 (0.2) & 64  & 73  & 89   & 99 & 123 & ---  \\
Teacher B2 (0.1) & 95 & 113   & --- & --- & --- & --- \\
Teacher B3 (0.05)  & 107   & --- & --- & --- & ---& ---\\
Teacher B4 (0.01) & --- & --- & --- & ---& --- & ---   \\
KD (5 Teachers) & 19  & 22  & 26  & 34    & 41    & 55  \\
Finetune  Teacher B1  & 35  & 38  & 59  & 72 & 78   & 108 \\
ResNet-110 & 80  & 80   & 80  & 80   & 80 & 80
\\
\bottomrule
\end{tabular}
\label{tb:diff_lr-epoch}
\end{minipage}

\vspace{-5mm}
\end{table*}

\section{Making Distillation Efficient}
\label{sec:kd-efficient}

\begin{table*}[ht]
 \begin{minipage}{0.32\textwidth}
 \makeatletter\def\@captype{table}\makeatother\caption{Accuracy at epoch 200 when distilling from teachers trained with different hyperparmeters (learning rates).\vspace{1mm}}
\begin{tabular}{lc} 
\toprule Model & Test Accuracy  \\
\midrule  All 5 teachers  & $72.5\%$ \\
Sample 1 Teacher & $72.1\%$ \\
Sample 2 Teachers & $72.4\%$ \\
Sample 3 Teachers  & $72.5\%$ \\
\bottomrule
\end{tabular}
\label{tb:diff_samp-acc}
 \end{minipage}
 \quad
  \begin{minipage}{0.66\textwidth}
\makeatletter\def\@captype{table}\makeatother\caption{Epochs necessary to reach the specified test accuracy when distilling from teachers trained with different hyperparameters (learning rates). \\ \phantom{blah blah blah blah blah blah blah blah} \vspace{1mm}}
\begin{tabular}{lcccccc}
\toprule
Model & $65.0\%$ & $66.0\%$ & $67.0\%$ & $68.0\%$ & $69.0\%$ & $70.0\%$ \\
\midrule  All 5 Teachers & 19  & 22  & 26  & 34    & 41    & 55 \\
Sample 1 Teacher & 26 & 34 & 37 & 44 & 58 & 72 \\
Sample 2 Teachers & 24 & 27 & 33 & 40 & 51 & 66 \\
Sample 3 Teachers & 22 & 23	& 30 & 37 & 48 & 60  \\
\bottomrule
\end{tabular}
\label{tb:diff_samp-epoch}
\end{minipage}
\vspace{-5mm}
\end{table*}

\begin{table}[htbp]
\vspace{-1mm}
\centering
\caption{Accuracy at epoch 200 when distilling for 20 epochs distributed at different times during training from teachers trained with different hyperparameters (learning rates). \vspace{1mm}}
\begin{tabular}{lc} 
\toprule Model & Test Accuracy  \\
\midrule 
Every Epoch (Baseline) &	$72.5\%$\\
\midrule
First 20 Epochs &	$69.3\%$\\
Middle 20 Epochs &	$68.9\%$\\
Last 20 Epochs &	$68.9\%$\\
First+Last 10 Epochs &	$69.8\%$\\
Every 10 Epochs &	$70.4\%$\\
2 Epochs Every 20 Epochs &	$70.8\%$\\
\bottomrule
\end{tabular}
\label{tb:diff_sched-acc}
\vspace{-7mm}
\end{table}

In the previous section, we showed that training future runs with KD provides a pareto improvement over training them from scratch, even when the teacher models are suboptimal and even taking into account the additional cost of KD.
However, that additional cost is significant.
Generally, performing a forward pass is about $1 \over 3$ of the total cost of a training step, meaning each additional teacher increases the cost of training by $1 \over 3$. Specifically, on P3.2xlarge instance, the time used for training a single ResNet56 on CIFAR100 without KD for one epoch is about 15s, adding a teacher would add about 5s. Therefore, performing KD with all 5 teachers as Section \ref{sec:kd_same} would increase the overall cost of training by 167\%.
In this section, we consider several approaches to reduce this overhead to further reduce the cost of training each subsequent model in a multi-run workflow like hyperparameter search.
In all experiments, we use teachers B0-B4 which are trained with different hyperparameters.

KD with multiple teachers is flexible, providing many different degrees of freedom to reduce costs.
We will look at two of these parts of the design space: only using some teachers on every step (\emph{sampling}) and performing distillation only during some parts of training (\emph{scheduling}).


\textbf{Sampling.}
We consider an orthogonal approach for reducing the cost of KD: using a randomly sampled subset of the available teachers on each step.
This reduces the overhead of distillation proportionally. By sampling 1 of 5 teachers, we increase the overall cost of training by only 33.3\%, comparing to 167\% when using 5 teachers for KD  every step.
Table \ref{tb:diff_samp-acc} and Table \ref{tb:diff_samp-epoch} show the results of doing so.
Sampling just one teacher on each step outperforms sampling a single fixed teacher (Table \ref{tb:fix_t-acc}) by 2.9 percentage points, despite the fact that many of the teachers sampled have suboptimal learning rates and accuracy.
It only sacrifices a small amount of accuracy compared to using all five teachers on every step (by 0.4 percentage points).
Although it learns about 30\% slower (Table \ref{tb:diff_samp-epoch}) than when using all five teachers, there is 5x less overhead from KD, more than making up the difference.

\textbf{Scheduling.}
Scheduling involves using KD only during specific parts of training.
During all other parts of training, we use a standard cross-entorpy loss between the student output and the ground-truth label.
For simplicity in this section, we fix the total number of KD epochs to 20 and study different ways to spend this budget during training.
By setting the total number of KD epochs to 20 and using five teachers, we increase the overall cost of training by only 17\%, compared to 167\% when performing KD at all steps.



We consider six different schedules for distributing these 20 epochs across training: the first 20 epochs, middle 20 epochs, last 20 epochs, first and last 10 epochs, 1 epoch for every 10 epochs, and 2 consecutive epochs for every 20 epochs.
The results of these experiments are in Table \ref{tb:diff_sched-acc}.
The best approaches involve performing distillation periodically, and doing so for longer blocks of 2 consecutive epochs performs better than shorter blocks of 1.
This approach outperforms performing distillation in one or two blocks during training by a margin of one or more percentage points.
None of these approaches match performing distillation on every step, but---considering the cost savings---the best approach is a pareto improvement.

\textbf{Summary.}
KD already provides a pareto improvement over training a new network from scratch, and the results in this section show that there are significant opportunities to make this process even more efficient by scheduling when to perform distillation and sampling teachers rather than using every teacher on every step.

\section{Conclusions}
\label{conclusion}

In this paper, we studied a common real-world scenario: training a sequence of related networks whose configurations vary in small ways.
This preparatory work---not the final training run---comprises the vast majority of the cost of developing models in practice.
To reduce this cost, we leveraged knowledge distillation from earlier runs to reduce the cost of future runs.
Not only did it prove effective, but it vastly reduced the number of steps necessary to reach the same accuracy as previous models, especially once we applied distillation in a targeted fashion.

This preliminary study serves as the basis for our deeper dive into this problem.
There are many remaining questions that must be answered before this strategy is ready for use in practice.
We intend to study how it works in larger-scale settings and how it applies to other topics areas like NLP.
Finally, we will evaluate the extent to which this strategy affects the trajectory and outcome of multi-run workloads like hyperparameter search.
However, these initial results have shed light on a promising future where model development costs are far lower than today.
\newpage

\bibliography{main}
\bibliographystyle{icml2022}


\newpage
\appendix
\section{KD with Good Teachers vs. Bad Teachers}

In Section \ref{sec:kd_same}, we studied the effect of performing KD using teachers that resulted from hyperparameter search over learning rates.
The final accuracies of these teachers varied widely, with several (B2, B3, and B4) performing worse than others.

In that experiment, we only showed results when distilling from all five teachers B0-B4.
However, each teacher may make different contributions to the performance of the student.
On the one hand, training with a worse-performing teacher may lead to a worse-performing student.
On the other hand, the literature shows \citep{allen2020towards} and our experiments have confirmed that student networks often outperform their teachers.
To address these questions, we compare distilling separately with the best-performing teacher (B1) and the worst-performing teacher (B4).


We apply 4 different KD strategies: only distilling from B1, only distilling from B4, sampling one network from among both on each step, and learning from both networks.
The results of this experiment are in Figure \ref{fig:good_bad}.
We find that learning from the worse-performing teacher (B4, light orange line) leads to a worse-performing student (dark orange line) compared to learning from the better-performing teacher (B1, light green line and dark green line).
The best performance occurs when learning from both teachers or sampling among them.

\section{Distilling Only When Teachers are Correct}

In this appendix, we study whether it is best to learn from teachers only when they are correct.
Teachers are not always correct; the argmax of their averaged output distributions may disagree with the label.
Is it important for the overall function that the student learns to receive even this incorrect signal?

\begin{figure}[ht]
\centering
\includegraphics[width=\linewidth]{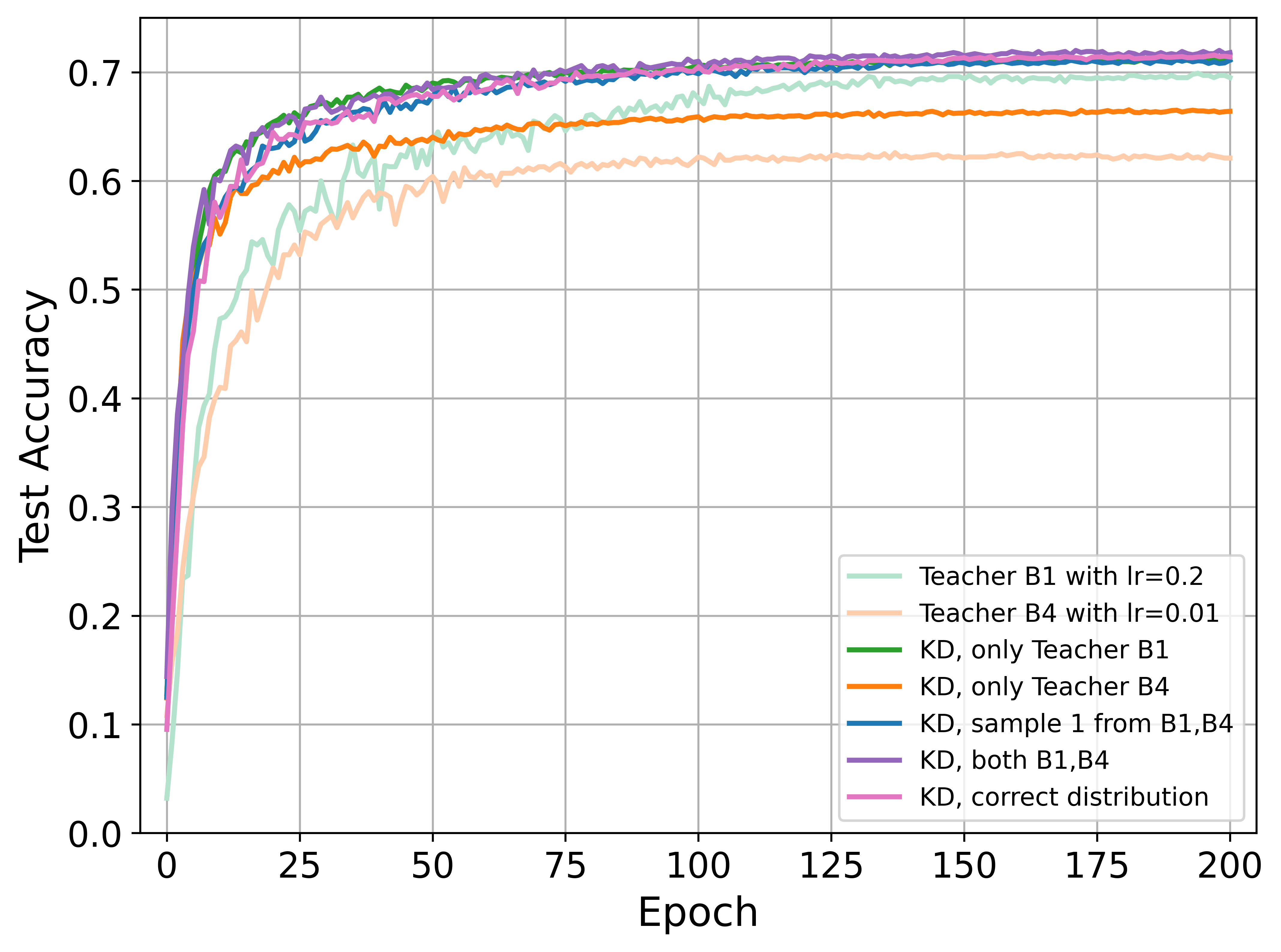}
  \caption{KD with various combinations of teachers B1 (the best performing teacher) and B4 (the worst performing teacher).}
\label{fig:good_bad}
\end{figure}

To address this question, we perform a modified form of KD where we discard the distillation part of the loss whenever the argmax of the averaged teacher output disagrees with the ground truth label.
The result of this experiment using teachers B1 and B4 is in Figure \ref{fig:good_bad} (pink line).
We find that it is detrimental to ignore the teachers when they are incorrect; doing so leads to accuracy that is 0.4 percentage points lower than learning from both teachers on every step (71.4\% accuracy vs. 71.8\% accuracy).
We conclude that even incorrect signals from the teachers are a useful part of the overall distillation process.


\section{Combining Scheduling and Sampling} 

In Section \ref{sec:kd-efficient}, we found that both scheduling and sampling could provide pareto improvements in the tradeoff between the cost of training student models and their final accuracy.
In this section, we combine these two methods.
We run KD using two different schedules (first and last 10 epochs, every 10 epochs), and we sample 1 teacher to use for KD on every batch.
Table \ref{tb:sched_samp-acc} shows the result of this experiment.
Interestingly, the results vary depending on the schedule chosen.
Sampling improves accuracy appreciably when using distillation at the beginning and end of training, whereas it hurts accuracy appreciably when using distillation for one epoch out of every ten epochs.
Due to Amdahl's Law, combining these two methods leads to diminishing returns in cost reduction (since they are both targeting the same source of cost: the overhead from performing distillation), so it is unclear whether it is beneficial to do so given these results.




\begin{table}[htbp]
\centering
\caption{Accuracy at epoch 200 when distilling from teachers trained with different hyperparameters when combining both the scheduling and sampling strategies.\vspace{1mm}}
\begin{tabular}{lc}
\toprule Model & Test Accuracy  \\
\midrule 
KD Every Epoch &	72.5\%\\
KD First+Last 10 Epochs, All 5 Teachers &	69.8\%\\
KD First+Last 10 Epochs, Sampling 1	& 70.3\%\\

KD Every 10 Epochs, All 5 Teachers &	70.4\%\\
KD Every 10 Epochs, Sampling 1 &	69.7\%\\
\bottomrule
\end{tabular}
\label{tb:sched_samp-acc}
\end{table}

\end{document}